\pdfoutput=1

\documentclass[11pt]{article}

\usepackage{acl}
\usepackage{amsfonts,amssymb}
\usepackage{times}
\usepackage{bbding}
\usepackage{latexsym}
\usepackage{bm}
\usepackage{enumitem}
\usepackage{hyperref}
\usepackage{makecell}
\usepackage[T1]{fontenc}
\usepackage{booktabs}
\usepackage[utf8]{inputenc}

\usepackage{microtype}
\usepackage{multirow}
\usepackage{multicol}
\usepackage{makecell}
\newcommand{\dashrule}[1][black]{%
  \color{#1}\rule[\dimexpr.2ex-.2pt]{4pt}{.4pt}\xleaders\hbox{\rule{2pt}{0pt}\rule[\dimexpr.2ex-.2pt]{4pt}{.4pt}}\hfill\kern0pt%
}
\usepackage{graphicx} 
\usepackage{amsmath}
\title{QURG: Question Rewriting Guided Context-Dependent \\Text-to-SQL Semantic Parsing}

\author{
  Linzheng Chai\textsuperscript{\rm 1},
  Dongling Xiao\textsuperscript{\rm 2},
  Jian Yang\textsuperscript{\rm 1},
  Liqun Yang\textsuperscript{\rm 1},\\
  {\bf Qian-Wen Zhang}\textsuperscript{\rm 2}, 
  {\bf Yunbo Cao}\textsuperscript{\rm 2},
  {\bf Zhoujun Li}\textsuperscript{\rm 1}, 
  {\bf Zhao Yan}\textsuperscript{\rm 2\thanks{\ Corresponding author.}},\\
  \textsuperscript{\rm 1}State Key Lab of Software Development Environment,
Beihang University, Beijing, China \\
  \textsuperscript{\rm 2}Tencent Cloud Xiaowei \\
  {dlxiao,cowenzhang,zhaoyan,yunbocao}@tencent.com \\
  {challenging,jiaya,lqyang,lizj}@buaa.edu.cn
}

\begin{document}
\maketitle
\begin{abstract}
Context-dependent Text-to-SQL aims to translate multi-turn natural language questions into SQL queries.
Despite various methods have exploited context-dependence information implicitly for contextual SQL parsing, there are few attempts to explicitly address the dependencies between current question and question context.
This paper presents QURG, a novel \underline{\textbf{QU}}estion \underline{\textbf{R}}ewriting \underline{\textbf{G}}uided approach to help the models achieve adequate contextual understanding.
Specifically, we first train a question rewriting model to complete the current question based on question context, and convert them into a rewriting edit matrix.
We further design a two-stream matrix encoder to jointly model the rewriting relations between question and context, and the schema linking relations between natural language and structured schema. Experimental results show that QURG significantly improves the performances on two large-scale context-dependent datasets SParC and CoSQL, especially for hard and long-turn questions.

\end{abstract}

\section{Introduction}

The past decade has witnessed increasing attention on text-to-SQL semantic parsing task, which aims to map natural language questions to SQL queries. Previously, works have mainly concentrated on the context-independent text-to-SQL task~\cite{zhongSeq2SQL2017,yu2018spider}, which translates single questions to SQL queries. The key to solving context-independent text-to-SQL is to model the relationships between questions and schema. 
Recent works have made great progress~\cite{wang2020rat,lin-bridging,cao2021lge} by employing schema linking mechanism which aligns schema items to entity references in the questions. 

\begin{figure}[t]
	  \centering
	  \setlength{\abovecaptionskip}{8pt}
	   \includegraphics[width=7.7cm]{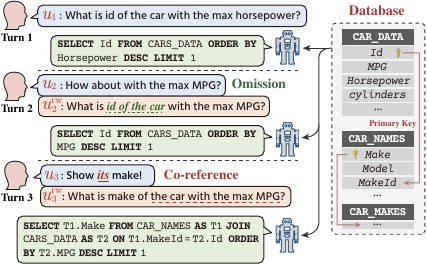}
	   \caption{An example of the context-dependent Text-to-SQL task with the phenomenon of co-reference and omission. $u_t^{\rm rw}$ denotes the rewritten question of the current question $u_t$ at $t$-th conversation turn.}\label{example}
\vspace{-10pt}
\end{figure}

With the extensive demand for interactive systems, the context-dependent text-to-SQL task which translates multi-turn questions to SQL queries has attracted more attention. Compared with the context-independent text-to-SQL task,  the context-dependent text-to-SQL task faces more challenges, that not only need to consider the relationship between natural language questions and the schema, but also the relationship between the current question and question context.
In a multi-turn scenario, as shown in Figure~\ref{example}, current questions may contain two contextual phenomena: \emph{co-reference} and \emph{omission} which are heavily associated with the historical context, meanwhile the question context may also contain information irrelevant to current questions. Thus models are required to selectively leverage contextual information to correctly address the user's intention of the current questions.

Previous works~\cite{hie-sql, yu2021score, scholak2021picard} on context-dependent text-to-SQL typically model the context dependencies in a simple way that feeds the concatenation of the current question, question context and schema into a neural networks encoder. To exploit context-dependence information,
\citet{hui2021dynamic} propose a dynamic relation decay mechanism to model the dynamic relationships between schema and question as conversation proceeds. Several works directly leverage historical generated SQL~\cite{zhang2019editing, hie-sql} or track interaction states associated with historical SQL~\cite{cai-wan-2020-igsql, istsql} to enhance the current SQL parsing.
Furthermore, \citet{yu2021score} and \citet{rat-sql-tc} introduce task-adaptive pre-trained language models and auxiliary training tasks on question and context to help models achieve adequate contextual understanding. 

However, these works neglect the explicit guidance on resolving contextual dependency, that SQL parsing and context understanding are coupled for model training.
The question rewriting (QR) task is to convert the current question into a self-contained question that can be understood without contextual information, and has been widely explored to represent multi-turn utterances currently \cite{su2019improving, pan2019improving, canard}. 
\citet{chen2021decoupled} is the first attempt at context-dependent text-to-SQL task by question rewriting, where a question rewrite model first explicitly completes the dialogue context, and then a context-independent Text-to-SQL parser follows. However, this approach relies on in-domain QR annotations and complex algorithms to obtain the rewritten question data.



\begin{figure}[t]
	  \centering
	  \setlength{\abovecaptionskip}{8pt}
	   \includegraphics[width=7.7cm]{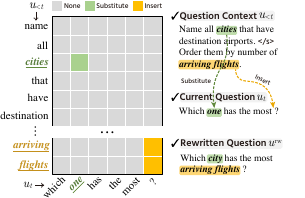}
	   \caption{An example of rewriting edit matrix. Given rewritten question $u^{\rm rw}_t$, we convert it to relations between current question $u_t$ and question context $u_{<t}$.}\label{rw_example}
\vspace{-10pt}
\end{figure}




To address the above limitations, we propose QURG, a novel \underline{QU}estion \underline{R}ewriting \underline{G}uided approach, which consists of three steps: 1) rewriting the current question into self-contained question and further converting it into a rewriting edit matrix; 2) jointly representing the rewriting matrix, multi-turn questions, and schema; 3) decoding the SQL queries.
Firstly, we train and evaluate the QR model on the out-of-domain dataset \textsc{Canard}~\cite{canard} and initialize the QR model with a pre-trained sequence generator for more precise rewritten questions. 
Secondly, inspired by~\cite{liu2020incomplete}, we propose to integrate rewritten results into the text-to-SQL task in the form of a rewriting relation matrix between question and context. 
We observed that directly replacing or concatenating original input with rewritten question may mislead the model for correctly SQL parsing. The reason is the unavoidable noise in rewritten questions and some questions are semantically complete and do not need to be rewritten.
The rewriting edit matrix denotes the relations between question and context words, these relations could clearly guide the model in solving long-range dependencies. 
Taking figure~\ref{rw_example} as an example, the rewritten question can be produced from original question through a series of edit operations in rewriting matrix (\textbf{\texttt{substitute}} \textit{“one”} in the current question with \textit{“cities”} in context, and \textbf{\texttt{insert}} \textit{“arriving flights”} before \textit{“?”}). 
Furthermore, we propose a two-stream relation matrix encoder based on the relation-aware Transformer (RAT)~\cite{shaw2018self} to jointly model the rewriting relation features between the current question and the context, and the schema linking relation features between multi-turn question and database schema. 
Finally, we aggregate the representations from the two relation matrix encoders to generate current SQL queries.

We evaluate our proposed QURG on two large-scale cross-domain context-dependent benchmarks: SParC~\cite{yu2019sparc} and CoSQL~\cite{yu2020cosql}. 
We summarize the contributions of this work as follows:\vspace{-0.25cm}
\begin{itemize}[leftmargin=*]
\item We present a novel context-dependent text-to-SQL framework QURG that explicitly guides models to resolve contextual dependencies.\vspace{-0.25cm}
\item Our framework incorporates rewritten questions in a novel way that explicitly represents multi-turn questions through rewriting relation matrix and two-stream relation matrix encoder.\vspace{-0.25cm}
\item Experimental results show that QURG achieves comparable performance to recent state-of-the-art works on two context-dependent text-to-SQL datasets. Besides, we further explore different approaches to incorporate rewritten questions into text-to-SQL to access the advantages of QURG.
\end{itemize}
 
\section{Related Work}
\subsection{Text-to-SQL} 
The text-to-SQL task aims to map natural language questions to database-related SQL queries. Spider~\cite{yu2018spider} is a widely evaluated cross-domain context-independent dataset and numerous works have shown that modeling the relation between question and schema can effectively improve performance on Spider.  \citet{shaw2018self} adopt relation-aware Transformer (RAT) ~\cite{shaw2018self} to encode the relational position for sentence representations, which has been widely transferred to text-to-SQL works~\cite{wang2020rat,lin-bridging,scholak2021duorat,yu2020grappa} to encode the schema-linking relations between natural language questions and structured database schema. \citet{cao2021lge} further improve relations modeling by line graph neural networks.

In face of the \emph{co-reference} and \emph{omission} in multi-turn questions, context-dependent text-to-SQL task is more challenging. Several works \cite{zhang2019editing, istsql, hie-sql} utilize previously generated SQL queries to resolve long-range dependency and improve the parsing accuracy. \citet{cai-wan-2020-igsql} and \citet{hui2021dynamic} use graph neural network to jointly encode multi-turn questions and schema. Inspired by the success of pre-trained models \cite{t5,bert,alm,ganlm,bart}, \citet{yu2021score} and \citet{rat-sql-tc} propose auxiliary state switch prediction tasks to model multi-turn question relations. \citet{scholak2021picard} simply constrain the auto-regressive decoders of super large pre-trained language models T5-3B. \citet{chen2021decoupled} propose a dual learning method to produce rewritten question data with in-domain QR annotations and directly use rewritten questions to generate SQL queries. In this work, we propose to adopt the rewriting matrix to explicitly model the relationships between the current question and context, and enhance the ability to capture long-range contextual dependencies.
\subsection{Question Rewriting (QR)}
\vspace{-2pt}
Question rewriting is to complete the \emph{co-reference} and \emph{omission} in the current question with historical context information, and help models understand multi-turn questions~\cite{elgohary2019can}. Most works \cite{pan2019improving,DBLP:rewrite_transformer,DBLP:rewrite_JD} conduct experiments on QR task as a sequence generation task with the copy mechanism. While \citet{liu2020incomplete} and \citet{hao2021rast} formulate the task as a semantic segmentation task and sequence-tagging task respectively.
Moreover, QR is widely applied to downstream tasks like conversational question answer (CQA) \cite{DBLP:pipeline_cqa,DBLP:naacl_cqa_2021,DBLP:excord,low_resource_template,kinet}, conversational retrieval \cite{DBLP:sigir_2021_cast19,yu_Few-Shot_rewrite} and context-dependent text-to-SQL \cite{chen2021decoupled}. Differently, we convert rewritten questions to rewriting relation matrix between question and context, and propose a two-stream relation matrix encoder to jointly model the rewriting and schema linking relations for context-dependent text-to-SQL parsing. Furthermore, we 
\vspace{-2pt}
\section{Preliminaries}
\vspace{-2pt}
In this section, we first formalize the context-dependent Text-to-SQL task, and then we introduce the relation-aware Transformer (RAT)~\cite{wang2020rat}, which is widely adopted to encode relations between sequence elements in text-to-SQL tasks, and which we use to build our two-stream encoder.\vspace{-5pt}
\subsection{Task Formulation}
\vspace{-2pt}
The context-dependent text-to-SQL task is to generate the SQL query $y_{t}$ given current user question $u_{t}$, historical question context $u_{<t}=\{u_1,u_2,$ $\ldots,u_{t-1}\}$ , and database schema $\mathcal{S}=\langle\mathcal{T},\mathcal{C}\rangle$, which consists of a series of tables $\mathcal{T} = \{t_{1},...,t_{|\mathcal{T}|}\}$ and columns $\mathcal{C}=\{c_{1},...,c_{|\mathcal{C}|}\}$. \vspace{-2pt}
\subsection{Relation-Aware Transformer (RAT)}
\vspace{-2pt}
The relation-aware transformer is an extension of the vanilla transformer~\cite{transformer}. RAT can integrate the pre-defined relation features by adding relation embedding to the self-attention mechanism of the vanilla transformer.

The vanilla Transformer is a model architecture which consists of a stack of multi-head self-attention layers, which has been widely used for tasks that process sequence inputs. Given input embedding sequence $\bm{{\rm X}}=\{\bm{{\rm x}}_i\}_{i=1}^n$ where $\bm{{\rm x}}_i\in\mathbb{R}^{d_x}$, each Transformer layer transform the input element $\bm{{\rm x}}_i$ into $\bm{{\rm y}}_i$ with $H$ heads as follows:\vspace{-4pt}
\begin{align}
e_{ij}^{(h)}\!\!&=\frac{\bm{{\rm x}}_i\bm{{\rm W}}_{Q}^{(h)}\left(\bm{{\rm x}}_j\bm{{\rm W}}_{K}^{(h)}\right)^{\top}}{\sqrt{d_{z}/H}}  \\
\alpha_{i j}^{(h)}\!\!&=\underset{j}{\operatorname{Softmax}}\left(e_{ij}^{(h)}\right) \\[-2mm]
\bm{{\rm z}}_i^{(h)}\!\!&=\sum_{j=1}^{n} \alpha_{ij}^{(h)}\left(\bm{{\rm x}}_j\bm{{\rm W}}_{V}^{(h)}\right)\\[-2mm]
\bm{{\rm z}}_i&=\operatorname{Concat} (\bm{{\rm z}}_i^{(1)}, ... ,\bm{{\rm z}}_i^{(H)}) \\
\widetilde{\bm{{\rm y}}}_i&=\operatorname{LayerNorm}(\bm{{\rm x}}_i+\bm{{\rm z}}_i) \\
\bm{{\rm y}}_i&=\operatorname{LayerNorm}(\widetilde{\bm{{\rm y}}}_i+\operatorname{FC}(\operatorname{ReLU}(\widetilde{\bm{{\rm y}}}_i)))
\end{align}

\vspace{-4pt}where $h$ denotes the $h$-th head, $a_{ij}^{(h)}$ is the attention weights, $\operatorname{Concat}(\cdot)$ is a concatenate operation, $\operatorname{FC(\cdot)}$ is a full-connected layer, $\operatorname{LayerNorm}(\cdot)$ is layer normalization, $\operatorname{ReLU(\cdot)}$ is the activation function and $\bm{{\rm W}}_Q^{(h)}$,$\bm{{\rm W}}_K^{(h)}$,$\bm{{\rm W}}_V^{(h)}$ are learnable projection parameters.

Compared to the vanilla transformer, RAT integrates learnable relation embedding into the self-attention module to bias model toward pre-defined relational information as:\vspace{-5pt}
\begin{align}
\!\!\!e_{i j}^{(h)}&=\frac{\bm{{\rm x}}_{i} \bm{{\rm W}}_{Q}^{(h)}\left(\bm{{\rm x}}_j \bm{{\rm W}}_{K}^{(h)}+\bm{{\rm r}}_{ij}^{K}\right)^{\top}}{\sqrt{d_{z} / H}} \\
\!\!\!\bm{{\rm z}}_{i}^{(h)}&=\sum_{j=1}^{n}\alpha_{ij}^{(h)}\left(\bm{{\rm x}}_j\bm{{\rm W}}_{V}^{(h)} +\bm{{\rm r}}_{ij}^{V}\right)^{\top}
\end{align}
where $\bm{{\rm r}}_{ij}$ is the pre-defined relation embedding between input elements $\bm{{\rm x}}_i$ and $\bm{{\rm x}}_j$.\vspace{-5pt}

\section{Methodology}
\begin{figure*}
	\centering
	\setlength{\abovecaptionskip}{7pt}
    \setlength{\belowcaptionskip}{-5pt}
	\includegraphics[width=16cm]{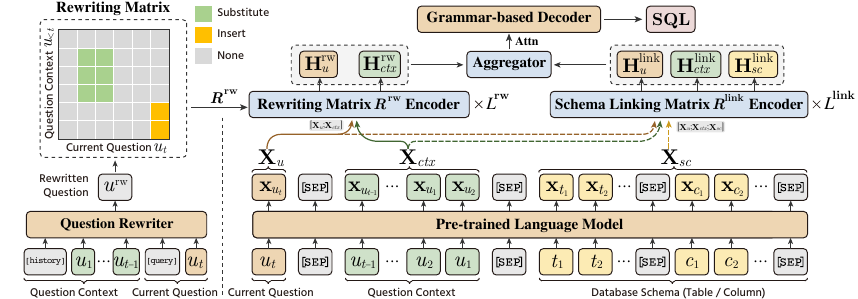}
	\caption{Illustration of QURG framework: \textbf{Left:} Question rewriting module which produces rewritten questions $u_t^{{\rm rw}}$ and rewriting matrix $R^{{\rm rw}}$. \textbf{Right:} PLM encoder and our proposed \textbf{two-stream relation matrix encoders}.}
	\label{fig_overview}
\end{figure*}

\vspace{-2pt}
Figure~\ref{fig_overview} illustrates our framework of QURG. It contains three parts: 1) Question rewriting model which is employed to obtain rewritten question $u_t^{\rm rw}$ from current question $u_t$ and context $u_{<t}$. 2) Rewriting matrix generator which converts rewritten question to word-level relation matrix between $u_t$ and $u_{<t}$. 3) SQL parser with two-stream encoder which can effectively integrate rewrite matrix for solving context dependencies.

\subsection{Question Rewriting Model}
Following \citet{lin2020conversational} and \citet{DBLP:excord}, we employ a pre-trained T5-base sequence generator~\cite{t5} as our QR model. Due to the lack of QR annotations on the text-to-SQL task, we directly use the \emph{out-of-domain} QR dataset~\textsc{Canard}~\cite{canard} for QR model training and evaluation. Specifically, given the current user question $u_t$ and historical context $u_{<t}$, we train QR models to produce rewritten question $u_t^{\rm rw}$ as: $\mathcal{P}(u_t^{\rm rw}|\{\texttt{[history]},u_{t-1},\texttt{[query]},u_t\})$, where \texttt{[history]} and \texttt{[query]} are special symbols to distinguish the context input and current question.

\subsection{Rewriting Matrix Construction}
Instead of feeding the rewritten question directly into the text-to-SQL model, we further convert the rewritten question into rewriting matrix which contains the key information to resolve context dependencies in current question. 
Following \citet{liu2020incomplete}, we adopt a heuristic method to construct the bi-directional rewriting matrix 
$R^{\rm rw}\in\mathbb{R}^{(|X_{\rm utter}|\times |X_{\rm utter}|)}$, where $X_{\rm utter}$ is the concatenate of historical context $u_{<t}$ and current question $u_t$. Firstly, we compare $u_t$ with $u_t^{\rm rw}$ to find the Longest Common Subsequence (LCS), for each word in $u_t^{\rm rw}$, if it is not in LCS, we will tag it as \textbf{\texttt{ADD}}. Similarly, for each word in $u_t$, if it is not in LCS, we tag it as \textbf{\texttt{DEL}}. After tagging, we merge consecutive words of the same tag to obtain \textbf{\texttt{ADD}} span and \textbf{\texttt{DEL}} span. Secondly, we traverse the \textbf{\texttt{ADD}} span in $u_t^{\rm rw}$, if it appears in the historical context $u_{<t}$ and corresponds to \textbf{\texttt{DEL}} in the same position in $u_t$, then we consider it as a “\textbf{\texttt{substitute}}” operation between original question $u_t$ and historical context $u_{<t}$, while there is no corresponding \textbf{\texttt{DEL}} in $u_t$, we consider it as a “\textbf{\texttt{insert}}” operation. Furthermore, if the \textbf{\texttt{ADD}} span does not appear in the context $u_{<t}$, we ignore it because these spans correspond to some unimportant word in most cases. 
Taking figure~\ref{rw_example} as an example, give current question $u_t$: \textit{“Which one has the most?”} and rewritten question $u^{\rm rw}_t$: \textit{“Which city has the most arriving flights?”}, the longest common sequence is \textit{“which has the most?”}. We tag \textit{“one”} in $u_t$ as \textbf{\texttt{DEL}}, and tag \textit{“city”} and \textit{“arriving flights”} in $u^{\rm rw}_t$ as \textbf{\texttt{ADD}}. Since the two \textbf{\texttt{ADD}} spans \textit{“city”} and \textit{“arriving flights”} all appear in context $u_{<t}$, the relation between \textit{“one”} in $u_t$ and \textit{“cities”} in $u_{<t}$ is “\textbf{\texttt{substitute}}”, the relation between \textit{“?”} in $u_t$ and \textit{“arriving flights"} in $u_{<t}$ is “\textbf{\texttt{insert}}”. Note that the rewriting matrix shown in Figure~\ref{rw_example} is uni-directional ($u_{<t}\rightarrow u_t$), when encoding later, question $u_t$ is concatenated with context $u_{<t}$, so we extend the uni-directional rewriting matrix to bi-directional rewriting matrix $R^{\rm rw}$ following the relation types between $u_t$ and $u_{<t}$ as shown in Table~\ref{rewrite_matrix_relation} (An example are described in Appendix~\ref{appendix_b}).
\begin{table}[t]
    \centering
    \setlength{\abovecaptionskip}{10pt}
\addtolength{\tabcolsep}{-1.5mm}
\resizebox{0.485\textwidth}{!}{
    \begin{tabular}{cccc}
    \toprule[1.0pt]
        Type of $x_i$&Type of $x_j$&Relation type&Description\\
    \midrule[0.3pt] 
        \multirow{3}{*}{\makecell[c]{Question\vspace{-0.1cm}\\$u_t$}} & \multirow{3}{*}{\makecell[c]{Context\vspace{-0.1cm}\\$u_{<t}$}} & \textsc{Q-C-Ins} & Insert $x_j$ before $x_i$\\ 
        && \textsc{Q-C-Sub}& Substitute $x_j$ for $x_i$\\ 
        && \textsc{None} & None operation\\ 
    \midrule[0.3pt] 
        \multirow{3}{*}{\makecell[c]{Context\vspace{-0.1cm}\\$u_{<t}$}} &\multirow{3}{*}{\makecell[c]{Question\vspace{-0.1cm}\\$u_t$}} & \textsc{C-Q-Ins} &Insert $x_i$ before $x_j$\\
        && \textsc{C-Q-Sub} &Substitute $x_i$ for $x_j$\\
        && \textsc{None} & None operation\\
    \bottomrule[1.0pt]
    \end{tabular}
    }
    \caption{Relation types between question and context. Relation edges exist from source token $x_i\in X_{\rm utter}$ to target token $x_j\in X_{\rm utter}$ if the pair meets one of the descriptions listed in the table, where $X_{\rm utter}$ is token set of $u_t$ and $u_{<t}$.}
    \label{rewrite_matrix_relation}
\end{table}

Through the above method, we can associate the existing omission and co-reference in the current question $u_t$ with the historical context $u_{<t}$, retaining context-dependencies in the form of a rewriting matrix, while ignoring trivial information or noise in the rewritten question $u_t^{\rm rw}$.\vspace{-5pt}
\subsection{\!\!\!QURG: SQL Parser with Rewriting Matrix}
Our QURG model is an extension of RAT-SQL \cite{wang2020rat} following the common \emph{encoder-decoder} architecture, which consists of three modules, as shown in Figure~\ref{fig_overview}: 1) Pre-trained Language Model (PLM) encoder which jointly transforms question $u_t$, context $u_{<t}$ and schema $\mathcal{S}$ into embedding as $\bm{{\rm X}}_u, \bm{{\rm X}}_{ctx}$ and $\bm{{\rm X}}_{sc}$ respectively; 2) Two-stream relation matrix input encoder which further encodes element embedding with pre-defined pairwise relation features as $\bm{{\rm H}}$; 3) Grammar-based decoder which generates SQL query corresponding to the current question.

\subsubsection{Pre-trained Language Model Encoder}
We concatenate the current question $u_t$, context $u_{<t}$ and schema $\mathcal{S}$ as the input sequence of pre-trained language models:
\begin{align*}
X\!=&\{\texttt{\![CLS]}\!,u_t,\texttt{[SEP]}\!,u_{t-1}, ..., u_1,\texttt{[SEP]}\!,t_1, \\ 
&,t_2,..., t_{|\mathcal{T}|},\texttt{[SEP]}\!,c_1,c_2,...c_{|\mathcal{C}|},\texttt{[SEP]}\!\}.
\end{align*}
Following \citet{cao2021lge}, we randomly shuffle the order of tables and columns in different mini-batches to alleviate the risk of over-fitting. Moreover, since each table name or column name may consist of multiple words, we use the average of the beginning and ending hidden vector as the schema element representation. Finally, the joint embedding vector of $X$ is represented as $\bm{{\rm X}}=\operatorname{Concat}(\bm{{\rm X}}_u;\bm{{\rm X}}_{ctx};\bm{{\rm X}}_{sc})$.

\subsubsection{Two-stream Relation Matrix Encoder}
This module contains two streams of relation matrix encoders: Schema Linking matrix $R^{\rm link}$ encoder and Rewriting matrix $R^{\rm rw}$ encoder. Firstly, the schema linking aids the model with aligning column/table references in the question and context to the corresponding schema columns/tables \texttt{id}. The schema linking relation matrix $R^{\rm link}$ is borrowed from RATSQL~\cite{wang2020rat} which builds relations between natural language and schema elements. Through the schema linking method, we can get schema linking matrix $R^{\rm link}\in\mathbb{R}^{(|X|\times|X|)}$. Then, the schema linking matrix $R^{\rm link}$ encoder takes joint embeddings of current question $\bm{{\rm X}}_{u}$, context $\bm{{\rm X}}_{ctx}$ and schema word $\bm{{\rm X}}_{sc}$ as input and applies $L^{\rm link}$ stacked RAT layers to produce contextual representation $\bm{{\rm H}}_u^{\rm link}$, $\bm{{\rm H}}_{ctx}^{\rm link}$ and $\bm{{\rm H}}_{sc}^{\rm link}$ respectively:\vspace{3pt}
\begin{align}
&\bm{{\rm H}}^{\rm link}_{(0)}=\operatorname{Concat}\left(\bm{{\rm X}}_u;\bm{{\rm X}}_{ctx};\bm{{\rm X}}_{sc}\right)\\
&\bm{{\rm H}}^{\rm link}_{(l)} = \operatorname{RAT}_{(l)}\left(\bm{{\rm H}}^{\rm link}_{(l-1)}, R^{\rm link}\right)
\end{align}

where $l\in [1, L^{\rm link}]$ denote the index of the $l$-th RAT layer.

Similarly, the rewriting matrix $R^{\rm rw}$ encoder takes the joint embeddings of the current question $\bm{{\rm X}}_u$ and context $\bm{{\rm X}}_{ctx}$ as input and applies $L^{\rm rw}$ stacked RAT layers to get the rewriting enhanced representations $\bm{{\rm H}}_u^{\rm rw}$, $\bm{{\rm H}}_{ctx}^{{\rm rw}}$ of question and context respectively:
\begin{align}
&\bm{{\rm H}}^{\rm rw}_{(0)} = \operatorname{Concat}\left(\bm{{\rm X}}_u;\bm{{\rm X}}_{ctx}\right)\\
&\bm{{\rm H}}^{\rm rw}_{(l)} = \operatorname{RAT}_{(l)}\left(\bm{{\rm H}}^{\rm rw}_{(l-1)}, R^{\rm rw}\right)
\end{align}

where $l\in[1, L^{\rm rw}]$ denote the index of the $l$-th RAT layer.

Finally, we aggregate the representations of the two-stream encoder as:\vspace{8pt}

$\bm{{\rm H}}\!=\!\operatorname{Concat}\!\left(\bm{{\rm H}}_u^{\rm link}\!+\!\bm{{\rm H}}_u^{\rm rw};\bm{{\rm H}}_{ctx}^{\rm link}\!+\!\bm{{\rm H}}_{ctx}^{\rm rw};\bm{{\rm H}}_{sc}^{\rm link}\right)$

\begin{table*}[t]
\centering
\setlength{\abovecaptionskip}{9pt}
\setlength{\belowcaptionskip}{-5pt}
\addtolength{\tabcolsep}{0mm}
\resizebox{1\textwidth}{!}{
\begin{tabular}{c|cccccccc}  
\toprule[1.0pt]
\multirow{2}{*}{\textbf{Dataset}} & Question & \multirow{2}{*}{Train / Dev / Test}&Database /&User&Average &\multirow{2}{*}{Vocab}& System &Cross  \\
&Interactions&&Domain&Questions&Turn&&Response&Domain   \\
\midrule[0.3pt]
SParC&4,298 &3,034 / 422 / 842 &200 / 138 &15,598&3.0&9,585&\XSolidBrush&\Checkmark\\
CoSQL&3,007 &2,164 / 293 / 551 &200 / 138 &12,726&5.2&3,794&\Checkmark&\Checkmark\\
\bottomrule[1.0pt]	
\end{tabular}
}
\caption{Detailed statistics for SParC dataset~\cite{yu2019sparc} and CoSQL dataset~\cite{yu2020cosql}.}
\label{tab_statistic_information}
\end{table*}

\subsubsection{Grammar-based Decoder}
We follow \citet{wang2020rat} and \citet{cao2021lge}, using a grammar-based syntactic neural decoder that generates the target SQL action sequence in the depth-first-search order of the abstract syntax tree (AST). We use single-layer LSTM~\cite{hochreiter1997long} as the auto-regressive decoder. At each step, the decoder predicts the probability of actions and uses the pointer mechanism to predict the probability of table/column \texttt{id}. We refer the reader to \cite{yin2018tranx} for details.


\section{Experiments}
In this section, we describe the experimental setups and evaluate the effectiveness of our proposed QURG. We compare QURG with previous works and conduct several ablation experiments. We also compare our method with the other two approaches of incorporating rewritten questions into text-to-SQL, to further verify the advantages of our QURG.
In addition, the experimental details of the QR model are described in  appendix~\ref{appendix_a}. 

\subsection{Experimental Setup}
\paragraph{Datasets} 

We train our QURG model on two large-scale cross-domain context-dependent text-to-SQL datasets, SparC~\cite{yu2019sparc} and CoSQL~\cite{yu2020cosql}. The details of those datasets are organized in Table~\ref{tab_statistic_information}.

\paragraph{Evaluation Metrics} For evaluation, we employ two main metrics on both SParC and CoSQL datasets: \textit{Question match} ($\bm{{\rm QM}}$) accuracy and \textit{Interaction match} ($\bm{{\rm IM}}$) accuracy. Specifically, for $\bm{{\rm QM}}$, if all clauses in a predicted SQL are exactly matching those of the target SQL, the matching score is $1.0$, otherwise, the score is $0.0$.  For $\bm{{\rm IM}}$, if all the predicted SQL in interaction is correct, the interaction match score is $1.0$, otherwise the score is $0.0$.

\paragraph{Implementation Details}
For Text-to-SQL tasks, we use \textsc{Electra}~\cite{clark2020electra} as our pre-trained language model for all experiments. We set the learning rate to $1e$-$4$, batch size to $32$, and the maximum gradient norm to $10$. The number of training epochs is $300$ and $320$ for SParC and CoSQL respectively. The numbers of RAT layers $L^{\rm link}=8$ for schema linking matrix encoder and $L^{\rm rw}=4$ for rewriting matrix encoder respectively. During inference, we set the beam size to $5$ for SQL parsing. Models are trained with $8$ NVIDIA V100 GPU cards.

\subsection{Experimental Results}
\begin{table}[hb]
\centering
\addtolength{\tabcolsep}{0mm}
\resizebox{0.45\textwidth}{!}{
\begin{tabular}{c|cc}  
\toprule[1.0pt]
\multirow{2}{*}{\textbf{Models}} &\multicolumn{2}{c}{\textbf{SParC}}\\
&$\bm{{\rm QM}}$ & $\bm{{\rm IM}}$ \\
\midrule[0.3pt]
EditSQL~\cite{zhang2019editing}&47.2&29.5\\
GAZP~\cite{zhong-etal-2020-grounded}&48.9&29.7\\
IGSQL~\cite{cai-wan-2020-igsql}&50.7&32.5\\
RichContext~\cite{2020How}&52.6&29.9\\
IST-SQL~\cite{istsql}& 47.6 & 29.9 \\
R$^2$SQL~\cite{hui2021dynamic}&54.1&35.2\\
DELTA~\cite{chen2021decoupled}&58.6&35.6\\
SCoRE$^{\flat}$~\cite{yu2021score}&62.2&42.5\\
HIE-SQL$^{\flat}$~\cite{hie-sql}&64.7&45.0\\
RAT-SQL+TC$^{\flat}$~\cite{rat-sql-tc}&64.1&44.1\\
\midrule[0.3pt]
\textbf{{QURG}~(Ours)}&\textbf{64.9}&\textbf{46.5}\\
\bottomrule[1.0pt]
\end{tabular}
}
\caption{Performances on the development set of SParC dataset. The models with $\!^\flat\!$ mark employ task adaptive pre-trained language models. } 
\label{sparc_result}
\end{table}

\begin{table*}[!h]
\centering
\setlength{\abovecaptionskip}{7pt}
\setlength{\belowcaptionskip}{-0.05cm}
\addtolength{\tabcolsep}{-0.1mm}
\resizebox{0.995\textwidth}{!}{
\begin{tabular}{c|ccccc|cccc} 
\toprule[1.0pt]
\textbf{SParC}~($\rightarrow$)&\textbf{Turn} $\bm{1}$&\textbf{Turn} $\bm{2}$&\textbf{Turn} $\bm{3}$&\!\textbf{Turn}$\geqslant\bm{4}$\!\!&\multirow{2}{*}{/}&$\bm{{\rm Easy}}$& $\!\bm{{\rm Medium}}$\!&$\bm{{\rm Hard}}$&$\bm{{\rm Extra}}$\\
\textbf{Models~($\downarrow$)}&\#$~422$&\#$~422$&\#$~270$&\#$~89$&&\#$~483$&\#$~441$&\#$~145$&\#$~134$\\
\midrule[0.3pt]
EditSQL$^{\;a}$&62.2&45.1&36.1&19.3&/&68.8&40.6&26.9&12.8\\
IGSQL$^{\;b}$&63.2&50.8&39.0&26.1&/&70.9&45.4&29.0&18.8\\
R$^2$SQL$^{\;c}$&67.7&55.3&45.7&33.0&/&75.5&51.5&35.2&21.8\\
RAT-SQL+TC$^{\;d}$&\textbf{75.4}&64.0&\textbf{54.4}&40.9&/&-&-&-&-\vspace{-6pt}\\
\multicolumn{10}{l}{\dashrule\vspace{-1pt}} \\
\textbf{QURG (Ours)} &\textbf{75.4}&\textbf{66.1}&53.7&\textbf{44.3}&/&\textbf{80.1}&\textbf{64.4}&\textbf{43.4}&\textbf{35.1}\\
\midrule[0.3pt]
\textbf{CoSQL}~($\rightarrow$)&\textbf{Turn} $\bm{1}$&\textbf{Turn} $\bm{2}$&\textbf{Turn} $\bm{3}$&\textbf{Turn} $\bm{4}$&\!\!\!\textbf{Turn} $\!>\!\bm{4}$&$\bm{{\rm Easy}}$& $\!\bm{{\rm Medium}}$\!&$\bm{{\rm Hard}}$&$\bm{{\rm Extra}}$\\
\textbf{Models~($\downarrow)$}&\#$~293$&\#$~285$&\#$~244$&\#$~114$&\#$~71$&\#$~417$&\#$~320$&\#$~163$&\#$~107$\\
\midrule[0.3pt]
EditSQL$^{\;a}$&50.0&36.7&34.8&43.0&23.9&62.7&29.4&22.8&9.3\\
IGSQL$^{\;b}$&53.1&42.6&39.3&43.0&31.0&66.3&35.6&26.4&10.3\\
IST-SQL$^{\;e}$&56.2& 41.0&41.0&41.2&26.8&66.0&36.2&27.8&10.3\\
SCoRE$^{\;f}$&60.8& 53.0&47.5&49.1&32.4&-&-&-&-\vspace{-6pt} \\
\multicolumn{10}{l}{\dashrule\vspace{-1pt}} \\
\textbf{QURG (Ours)}&\textbf{64.5}&\textbf{55.4}&\textbf{55.7}&\textbf{50.0}&\textbf{42.3}&\textbf{77.2}&\textbf{50.0}&\textbf{40.5}&\textbf{20.6}\\
\bottomrule[1.0pt]
\end{tabular}
}
\caption{Detailed question match accuracy~($\bm{{\rm QM}}$)~results in different interaction turns and SQL difficulties on the development set of SParC and CoSQL datasets. Results of $^{a\;}$\protect\cite{zhang2019editing},$^{b\;}$\protect\cite{cai-wan-2020-igsql},$^{c\;}$\protect\cite{istsql},$^{d\;}$\protect\cite{rat-sql-tc},$^{e\;}$\protect\cite{istsql} and $^{f\;}$\protect\cite{yu2021score} are from the original paper.}
\label{tab_turns}
\end{table*}

\begin{table}[t]
\centering
\setlength{\abovecaptionskip}{5pt}
\setlength{\belowcaptionskip}{-7pt}
\addtolength{\tabcolsep}{0mm}
\resizebox{0.45\textwidth}{!}{
\begin{tabular}{c|cc}  
\toprule[1.0pt]
\multirow{2}{*}{\textbf{Models}} &\multicolumn{2}{c}{\textbf{CoSQL}}\\
&$\bm{{\rm QM}}$ & $\bm{{\rm IM}}$ \\
\midrule[0.3pt]
EditSQL~\cite{zhang2019editing}&39.9&12.3\\
GAZP~\cite{zhong-etal-2020-grounded}&42.0&12.3\\
IGSQL~\cite{cai-wan-2020-igsql}&44.1&15.8\\
RichContext~\cite{2020How}&41.0&14.0\\
IST-SQL~\cite{istsql}& 44.4 & 14.7 \\
R$^2$SQL~\cite{hui2021dynamic}&45.7&19.5\\
DELTA~\cite{chen2021decoupled}&51.7&21.5\\
SCoRE$^{\flat}$~\cite{yu2021score}&52.1&22.0\\
\textsc{Picard}$^{\dag}$~\cite{scholak2021picard}&\textbf{56.9}&24.2\\
HIE-SQL$^{\flat}$~\cite{hie-sql}&56.4&\textbf{28.7}\\
\midrule[0.3pt]
\textbf{{QURG}~(Ours)}&56.6&26.6\\
\bottomrule[1.0pt]
\end{tabular}
}
\caption{Performances on the development set of CoSQL dataset. Model with $^\dag$ mark is based on the super large T5-$3$B~\protect\cite{t5} pre-trained model.} 
\label{cosql_result}
\end{table}


As shown in Table~\ref{sparc_result} and~\ref{cosql_result}, we compare the performances of QURG with previous works on the development set of SParC and CoSQL datasets. QURG achieves comparable performance to previous state-of-the-art methods, including SCoRE~\cite{yu2021score}, RAT-SQL-TC~\cite{rat-sql-tc} and HIE-SQL~\cite{hie-sql} which effectively promote performances by using task-adaptive pre-trained language models. Besides, our QURG outperforms DELTA~\cite{chen2021decoupled} which directly uses rewritten questions to predict SQL queries. In terms of $\bm{{\rm IM}}$ accuracy on the CoSQL, QURG also surpasses \textsc{Picard}~\cite{scholak2021picard} which is based on the super large pre-trained models T5-3B~\cite{t5}.\vspace{-2pt}

To further study the advantages of QURG on contextual understanding, as shown in Table~\ref{tab_turns}, we evaluate the performances of the different question turns on SparC and CoSQL, and compare our QURG with previous powerful methods. As the number of turns increases, the difficulty of the task increases because models need to resolve the co-reference and omission with longer dependencies. Besides, our QURG can achieve more improvements as the interaction turn increase. Furthermore, we evaluate the performance of QURG on the different difficulty levels of target SQL as shown in the right of Table~\ref{tab_turns}, we observe that our QURG consistently achieves comparable performances to recent state-of-the-art works.\vspace{-3pt}

\begin{table}[t]
\centering
\setlength{\abovecaptionskip}{5pt}
\setlength{\belowcaptionskip}{-7pt}
\addtolength{\tabcolsep}{0mm}
\resizebox{0.39\textwidth}{!}{
\begin{tabular}{ll|cccc} 
\toprule[1.0pt]
\multirow{2}{*}{\!\!\texttt{\!\!}}&\multirow{2}{*}{\!\!\textbf{Models}}&\multicolumn{2}{c}{\textbf{SParC}}&\multicolumn{2}{c}{\textbf{CoSQL}}\\
&&$\bm{{\rm QM}}$&$\bm{{\rm IM}}$&$\bm{{\rm QM}}$&$\bm{{\rm IM}}$\\
\midrule[0.3pt]
\texttt{\!\!\!\!}&\!\!\textbf{QURG}&\textbf{64.9}&\textbf{46.5}&\textbf{56.6}&\textbf{26.6}\\
\texttt{\!\!\!\!}&$-{R^{\rm rw}}$&62.6&43.6&55.6&25.2\\
\texttt{\!\!\!\!}&$-{\rm Enc^{\rm rw}}$&63.4&44.7&55.0&24.5\\
\bottomrule[1.0pt]
\end{tabular}
}
\caption{Ablation studies for the components of QURG. Note that ${\rm-Enc_{\rm rw}}$ is also the baseline without the integration of rewritten questions $u_{<t}$.}
\label{tab_qurg_ablation}
\end{table}

\subsection {Ablation Study}
As shown in Table ~\ref{tab_qurg_ablation}, we conduct several ablation studies to evaluate the contribution of rewriting matrix integration for our QURG. 

\begin{figure*}[t]
	\centering
	\setlength{\abovecaptionskip}{5pt}
    \setlength{\belowcaptionskip}{-5pt}
	\includegraphics[width=16cm]{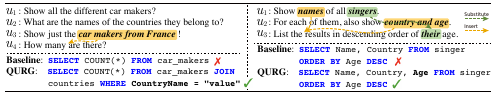}
	\caption{Case studies on SParC dataset. \textbf{Baseline} denotes the QURG model without rewriting matrix encoder. }
	\label{case_study}
\end{figure*}

To explore the effects of the rewriting matrix ($-R^{\rm rw}$), we set all the relation types in the rewriting matrix to \textsc{None} and keep the model structure unchanged. It degrades the model performances on both datasets by $1.0\%$-$2.9\%$ which confirms that rewriting matrix can effectively improve SQL parsing ability through enhanced context understanding. Then we further remove the whole rewriting matrix encoder ($-{\rm Enc}^{\rm rw}$) to verify the effect of the additional encoder parameters on question $u_t$ and context $u_{<t}$, we observe that the additional parameters slightly degrade the performance on SParC, while slightly improving on CoSQL ($-{\rm Enc}^{\rm rw}\rightarrow-R^{\rm rw}$), which indicates that the additional parameters have little effect on the improvements of QURG. 


\begin{table}[t]
\centering
\setlength{\abovecaptionskip}{5pt}
\resizebox{0.35\textwidth}{!}{
\begin{tabular}{l|cccc} 
\toprule[1.0pt]
\multirow{2}{*}{\!\!\textbf{Models}}&\multicolumn{2}{c}{\textbf{SParC}}&\multicolumn{2}{c}{\textbf{CoSQL}}\\
&${\rm QM}$&${\rm IM}$&${\rm QM}$&${\rm IM}$\\
\midrule[0.3pt]
\!\!\textsc{Only}&52.4&29.4&45.9&15.0\\
\!\!\textsc{Concat}&62.3&41.2&53.0&20.8\\
\!\!\textbf{QURG}&\textbf{64.9}&\textbf{46.5}&\textbf{56.6}&\textbf{26.6}\\
\bottomrule[1.0pt]
\end{tabular}
}
\caption{Studies on different approaches to inject rewritten question into context-dependent text-to-SQL.}
\label{tab_diff_methods}
\end{table}

\begin{figure}[t]
	  \centering
	  \setlength{\abovecaptionskip}{5pt}
	   \includegraphics[width=7.7cm]{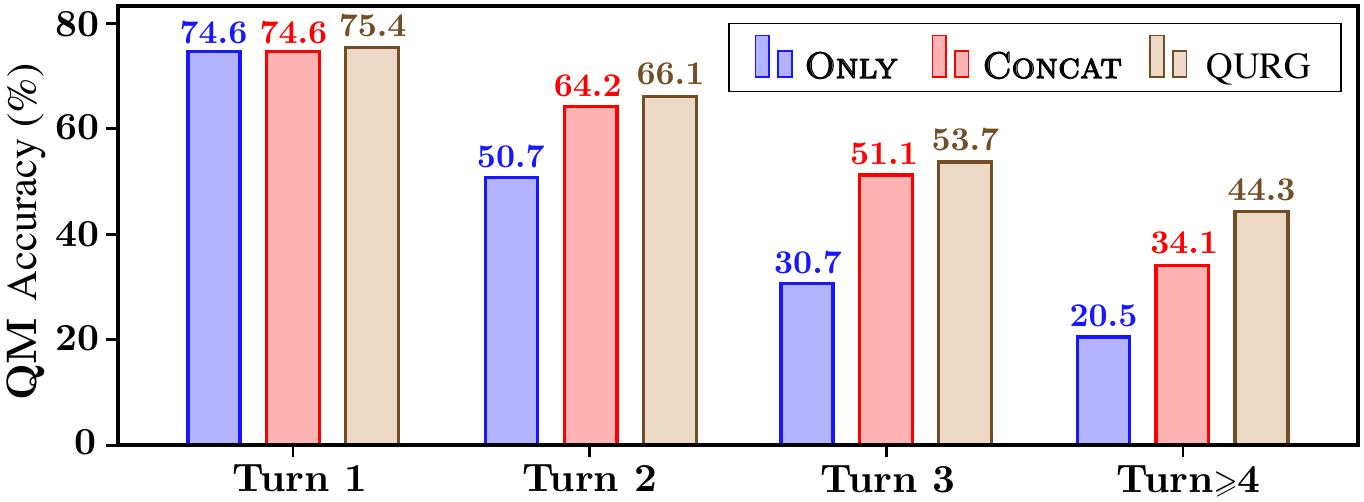}
	   \caption{Detailed results on different question turns for models \textsc{Only}, \textsc{Concat} and QURG.}\label{fig_diff_methods}
\end{figure}

Moreover, we explore the effects of different approaches to integrating rewritten questions into text-to-SQL tasks, as shown in Table~\ref{tab_diff_methods}: 1) “\textbf{\textsc{Only}}” indicates only using rewritten questions $u_t^{\rm rw}$ to generate SQL queries, discarding the original question $u_t$ and context $u_{<t}$; 2) “\textbf{\textsc{Concat}}” indicates concatenating original question $u_t$, context $u_{<t}$ with rewrite questions $u_t^{\rm rw}$, treating $u_t^{\rm rw}$ as additional information. 
As shown in Table~\ref{tab_diff_methods} and Figure~\ref{fig_diff_methods}, \textsc{Only} feeding rewritten questions into text-to-SQL models results in a substantial performance drop, since the QR model is trained on \emph{out-of-domain} data, the rewritten questions may contain a lot of noise.
For “\textsc{Concat}”, question and context are retained to meet the potential noise in rewritten questions, while it is still not ideal and causes performance to degraded against QURG, especially with the increase of turns, the performance of \textsc{Concat} decreases more significantly. Not all questions need to be rewritten, for the questions without co-references or omissions, \textsc{Concat} will introduce redundant information or noise.

Compared with the \textsc{Only} and \textsc{Concat}, QURG is more competitive by leveraging the rewriting matrix, which effectively preserves the rewriting information without introducing redundant information (if the question does not need to be rewritten, the rewriting matrix degrades into nondistinctive bias). 

\subsection{Case Study}
In figure \ref{case_study} , we offer some cases from the development set of SparC, in which QURG generates SQL queries correctly, while baseline (without rewriting matrix encoder $-{\rm Enc^{rw}}$) fails. In the first case,  the “\emph{car makers from France}” is omitted in the current question, the baseline model resolves the omitted subject “\emph{car maker}” while ignoring the condition “\emph{from France}”, our QURG utilizes the complete “\texttt{Insert}” relation between $u_4$ and context $u_{<4}$ to understand $u_4$ and generated correct SQL query. In the second case, “\emph{results}” in current question $u_3$ refers to “\emph{names, country and ages}” in context $u_{<3}$, “\emph{their}” refers to “\emph{singers}” in $u_1$. The baseline model failed to build the relation between “\emph{results}” and “\emph{age}”, and our QURG successfully incorporates this relation to model and products correct SQL query.

\section{Conclusions}
We propose QURG, a novel context-dependent text-to-SQL framework that utilizes question rewriting to resolve long-distance dependencies between the current question and historical context. Firstly, QURG adopts a pre-trained sequence generator to produce rewritten questions and further converts them to the rewriting relation matrix between question and context. Secondly, QURG employs a two-stream matrix encoder to incorporate the rewriting edit relations into a text-to-SQL model to enhance contextual understanding. Experimental results show that our QURG achieves comparable performance with recent state-of-the-art works on two large-scale context-dependent text-to-SQL datasets SParC and CoSQL.\vspace{-5pt}

\bibliography{anthology,custom}

\begin{thebibliography}{46}
\expandafter\ifx\csname natexlab\endcsname\relax\def\natexlab#1{#1}\fi

\bibitem[{Anantha et~al.(2021)Anantha, Vakulenko, Tu, Longpre, Pulman, and
  Chappidi}]{DBLP:naacl_cqa_2021}
Raviteja Anantha, Svitlana Vakulenko, Zhucheng Tu, Shayne Longpre, Stephen
  Pulman, and Srinivas Chappidi. 2021.
\newblock \href {https://arxiv.org/abs/2010.04898} {Open-domain question
  answering goes conversational via question rewriting}.
\newblock In \emph{ACL}.

\bibitem[{Bai et~al.(2023)Bai, Yang, Yang, Guo, and Li}]{kinet}
Jiaqi Bai, Ze~Yang, Jian Yang, Hongcheng Guo, and Zhoujun Li. 2023.
\newblock \href {https://doi.org/10.1109/TASLP.2023.3240654} {Kinet:
  Incorporating relevant facts into knowledge-grounded dialog generation}.
\newblock \emph{{IEEE} {ACM} Trans. Audio Speech Lang. Process.},
  31:1213--1222.

\bibitem[{Cai and Wan(2020)}]{cai-wan-2020-igsql}
Yitao Cai and Xiaojun Wan. 2020.
\newblock \href {https://doi.org/10.18653/v1/2020.emnlp-main.560} {{IGSQL}:
  Database schema interaction graph based neural model for context-dependent
  text-to-{SQL} generation}.
\newblock In \emph{Proceedings of the 2020 Conference on Empirical Methods in
  Natural Language Processing (EMNLP)}, pages 6903--6912, Online. Association
  for Computational Linguistics.

\bibitem[{Cao et~al.(2021)Cao, Chen, Chen, Zhao, Zhu, and Yu}]{cao2021lge}
Ruisheng Cao, Lu~Chen, Zhi Chen, Yanbin Zhao, Su~Zhu, and Kai Yu. 2021.
\newblock \href {https://doi.org/10.18653/v1/2021.acl-long.198} {{LGESQL}: Line
  graph enhanced text-to-{SQL} model with mixed local and non-local relations}.
\newblock In \emph{Proceedings of the 59th Annual Meeting of the Association
  for Computational Linguistics and the 11th International Joint Conference on
  Natural Language Processing (Volume 1: Long Papers)}, pages 2541--2555,
  Online. Association for Computational Linguistics.

\bibitem[{Chen et~al.(2021)Chen, Chen, Li, Cao, Ma, Wu, and
  Yu}]{chen2021decoupled}
Zhi Chen, Lu~Chen, Hanqi Li, Ruisheng Cao, Da~Ma, Mengyue Wu, and Kai Yu. 2021.
\newblock \href {https://doi.org/10.18653/v1/2021.findings-acl.270} {Decoupled
  dialogue modeling and semantic parsing for multi-turn text-to-{SQL}}.
\newblock In \emph{Findings of the Association for Computational Linguistics:
  ACL-IJCNLP 2021}, pages 3063--3074, Online. Association for Computational
  Linguistics.

\bibitem[{Choi et~al.(2018)Choi, He, Iyyer, Yatskar, Yih, Choi, Liang, and
  Zettlemoyer}]{quac}
Eunsol Choi, He~He, Mohit Iyyer, Mark Yatskar, Wen-tau Yih, Yejin Choi, Percy
  Liang, and Luke Zettlemoyer. 2018.
\newblock \href {https://doi.org/10.18653/v1/D18-1241} {{Q}u{AC}: Question
  answering in context}.
\newblock In \emph{Proceedings of the 2018 Conference on Empirical Methods in
  Natural Language Processing}, pages 2174--2184, Brussels, Belgium.
  Association for Computational Linguistics.

\bibitem[{Clark et~al.(2020)Clark, Luong, Le, and Manning}]{clark2020electra}
Kevin Clark, Minh-Thang Luong, Quoc~V. Le, and Christopher~D. Manning. 2020.
\newblock \href {https://openreview.net/forum?id=r1xMH1BtvB} {Electra:
  Pre-training text encoders as discriminators rather than generators}.
\newblock In \emph{International Conference on Learning Representations}.

\bibitem[{Dalton et~al.(2020)Dalton, Xiong, Kumar, and
  Callan}]{DBLP:sigir_2021_cast19}
Jeffrey Dalton, Chenyan Xiong, Vaibhav Kumar, and Jamie Callan. 2020.
\newblock \href {https://www.cs.cmu.edu/~callan/Papers/sigir20-dalton.pdf}
  {Cast-19: A dataset for conversational information seeking}.
\newblock In \emph{Proceedings of the 43rd International ACM SIGIR Conference
  on Research and Development in Information Retrieval}, pages 1985--1988.

\bibitem[{Elgohary et~al.(2019{\natexlab{a}})Elgohary, Peskov, and
  Boyd-Graber}]{canard}
Ahmed Elgohary, Denis Peskov, and Jordan Boyd-Graber. 2019{\natexlab{a}}.
\newblock \href {https://doi.org/10.18653/v1/D19-1605} {Can you unpack that?
  learning to rewrite questions-in-context}.
\newblock In \emph{Proceedings of the 2019 Conference on Empirical Methods in
  Natural Language Processing and the 9th International Joint Conference on
  Natural Language Processing (EMNLP-IJCNLP)}, pages 5918--5924, Hong Kong,
  China. Association for Computational Linguistics.

\bibitem[{Elgohary et~al.(2019{\natexlab{b}})Elgohary, Peskov, and
  Boyd-Graber}]{elgohary2019can}
Ahmed Elgohary, Denis Peskov, and Jordan Boyd-Graber. 2019{\natexlab{b}}.
\newblock Can you unpack that? learning to rewrite questions-in-context.
\newblock In \emph{Proceedings of the 2019 Conference on Empirical Methods in
  Natural Language Processing and the 9th International Joint Conference on
  Natural Language Processing (EMNLP-IJCNLP)}, pages 5918--5924.

\bibitem[{Hao et~al.(2021)Hao, Song, Wang, Xu, Tu, and Yu}]{hao2021rast}
Jie Hao, Linfeng Song, Liwei Wang, Kun Xu, Zhaopeng Tu, and Dong Yu. 2021.
\newblock \href {https://aclanthology.org/2021.emnlp-main.402/} {Rast:
  Domain-robust dialogue rewriting as sequence tagging}.
\newblock In \emph{Proceedings of the 2021 Conference on Empirical Methods in
  Natural Language Processing}, pages 4913--4924.

\bibitem[{Hochreiter and Schmidhuber(1997)}]{hochreiter1997long}
Sepp Hochreiter and J{\"u}rgen Schmidhuber. 1997.
\newblock \href
  {https://direct.mit.edu/neco/article-abstract/9/8/1735/6109/Long-Short-Term-Memory?redirectedFrom=fulltext}
  {Long short-term memory}.
\newblock \emph{Neural computation}, 9(8):1735--1780.

\bibitem[{Hui et~al.(2021)Hui, Geng, Ren, Li, Li, Sun, Huang, Si, Zhu, and
  Zhu}]{hui2021dynamic}
Binyuan Hui, Ruiying Geng, Qiyu Ren, Binhua Li, Yongbin Li, Jian Sun, Fei
  Huang, Luo Si, Pengfei Zhu, and Xiaodan Zhu. 2021.
\newblock \href {https://arxiv.org/abs/2101.01686} {Dynamic hybrid relation
  exploration network for cross-domain context-dependent semantic parsing}.
\newblock In \emph{Proceedings of the AAAI Conference on Artificial
  Intelligence}, volume~35, pages 13116--13124.

\bibitem[{Kim et~al.(2021)Kim, Kim, Park, and Kang}]{DBLP:excord}
Gangwoo Kim, Hyunjae Kim, Jungsoo Park, and Jaewoo Kang. 2021.
\newblock Learn to resolve conversational dependency: {A} consistency training
  framework for conversational question answering.
\newblock In \emph{ACL}.

\bibitem[{Lewis et~al.(2020)Lewis, Liu, Goyal, Ghazvininejad, Mohamed, Levy,
  Stoyanov, and Zettlemoyer}]{bart}
Mike Lewis, Yinhan Liu, Naman Goyal, Marjan Ghazvininejad, Abdelrahman Mohamed,
  Omer Levy, Veselin Stoyanov, and Luke Zettlemoyer. 2020.
\newblock {BART:} denoising sequence-to-sequence pre-training for natural
  language generation, translation, and comprehension.
\newblock In \emph{ACL 2020}, pages 7871--7880.

\bibitem[{Li et~al.(2021)Li, Zhang, Li, Wang, Wu, and Zhang}]{rat-sql-tc}
Yuntao Li, Hanchu Zhang, Yutian Li, Sirui Wang, Wei Wu, and Yan Zhang. 2021.
\newblock \href {https://arxiv.org/pdf/2112.08735.pdf} {Pay more attention to
  history: A context modeling strategy for conversational text-to-sql}.
\newblock \emph{arXiv:2112.08735}.

\bibitem[{Lin et~al.(2020{\natexlab{a}})Lin, Yang, Nogueira, Tsai, Wang, and
  Lin}]{lin2020conversational}
Sheng-Chieh Lin, Jheng-Hong Yang, Rodrigo Nogueira, Ming-Feng Tsai, Chuan-Ju
  Wang, and Jimmy Lin. 2020{\natexlab{a}}.
\newblock \href {https://arxiv.org/abs/2004.01909} {Conversational question
  reformulation via sequence-to-sequence architectures and pretrained language
  models}.
\newblock \emph{arXiv preprint arXiv:2004.01909}.

\bibitem[{Lin et~al.(2020{\natexlab{b}})Lin, Socher, and Xiong}]{lin-bridging}
Xi~Victoria Lin, Richard Socher, and Caiming Xiong. 2020{\natexlab{b}}.
\newblock \href {https://doi.org/10.18653/v1/2020.findings-emnlp.438} {Bridging
  textual and tabular data for cross-domain text-to-{SQL} semantic parsing}.
\newblock In \emph{Findings of the Association for Computational Linguistics:
  EMNLP 2020}, pages 4870--4888, Online. Association for Computational
  Linguistics.

\bibitem[{Liu et~al.(2021)Liu, Chen, Wu, He, and Zhou}]{DBLP:rewrite_JD}
Hang Liu, Meng Chen, Youzheng Wu, Xiaodong He, and Bowen Zhou. 2021.
\newblock \href {https://arxiv.org/abs/2102.04708} {Conversational query
  rewriting with self-supervised learning}.
\newblock In \emph{ICASSP}.

\bibitem[{Liu et~al.(2020{\natexlab{a}})Liu, Chen, Guo, Lou, Zhou, and
  Zhang}]{2020How}
Qian Liu, Bei Chen, Jiaqi Guo, Jian-Guang Lou, Bin Zhou, and Dongmei Zhang.
  2020{\natexlab{a}}.
\newblock \href {https://doi.org/10.24963/ijcai.2020/495} {How far are we from
  effective context modeling? an exploratory study on semantic parsing in
  context}.
\newblock In \emph{Proceedings of the Twenty-Ninth International Joint
  Conference on Artificial Intelligence, {IJCAI-20}}, pages 3580--3586.
  International Joint Conferences on Artificial Intelligence Organization.

\bibitem[{Liu et~al.(2020{\natexlab{b}})Liu, Chen, Lou, Zhou, and
  Zhang}]{liu2020incomplete}
Qian Liu, Bei Chen, Jian-Guang Lou, Bin Zhou, and Dongmei Zhang.
  2020{\natexlab{b}}.
\newblock \href {https://arxiv.org/abs/2009.13166} {Incomplete utterance
  rewriting as semantic segmentation}.
\newblock In \emph{Proceedings of the 2020 Conference on Empirical Methods in
  Natural Language Processing (EMNLP)}, pages 2846--2857.

\bibitem[{Pan et~al.(2019)Pan, Bai, Wang, Zhou, and Liu}]{pan2019improving}
Zhufeng Pan, Kun Bai, Yan Wang, Lianqiang Zhou, and Xiaojiang Liu. 2019.
\newblock \href {https://aclanthology.org/D19-1191/} {Improving open-domain
  dialogue systems via multi-turn incomplete utterance restoration}.
\newblock In \emph{Proceedings of the 2019 Conference on Empirical Methods in
  Natural Language Processing and the 9th International Joint Conference on
  Natural Language Processing (EMNLP-IJCNLP)}, pages 1824--1833.

\bibitem[{Raffel et~al.(2020)Raffel, Shazeer, Roberts, Lee, Narang, Matena,
  Zhou, Li, and Liu}]{t5}
Colin Raffel, Noam Shazeer, Adam Roberts, Katherine Lee, Sharan Narang, Michael
  Matena, Yanqi Zhou, Wei Li, and Peter~J. Liu. 2020.
\newblock \href {http://jmlr.org/papers/v21/20-074.html} {Exploring the limits
  of transfer learning with a unified text-to-text transformer}.
\newblock \emph{Journal of Machine Learning Research}, 21(140):1--67.

\bibitem[{Scholak et~al.(2021{\natexlab{a}})Scholak, Li, Bahdanau, de~Vries,
  and Pal}]{scholak2021duorat}
Torsten Scholak, Raymond Li, Dzmitry Bahdanau, Harm de~Vries, and Christopher
  Pal. 2021{\natexlab{a}}.
\newblock \href {https://arxiv.org/abs/2010.11119} {Duorat: Towards simpler
  text-to-sql models}.
\newblock In \emph{Proceedings of the 2021 Conference of the North American
  Chapter of the Association for Computational Linguistics: Human Language
  Technologies}, pages 1313--1321.

\bibitem[{Scholak et~al.(2021{\natexlab{b}})Scholak, Schucher, and
  Bahdanau}]{scholak2021picard}
Torsten Scholak, Nathan Schucher, and Dzmitry Bahdanau. 2021{\natexlab{b}}.
\newblock \href {https://doi.org/10.18653/v1/2021.emnlp-main.779} {{PICARD}:
  Parsing incrementally for constrained auto-regressive decoding from language
  models}.
\newblock In \emph{Proceedings of the 2021 Conference on Empirical Methods in
  Natural Language Processing}, pages 9895--9901, Online and Punta Cana,
  Dominican Republic. Association for Computational Linguistics.

\bibitem[{Shaw et~al.(2018)Shaw, Uszkoreit, and Vaswani}]{shaw2018self}
Peter Shaw, Jakob Uszkoreit, and Ashish Vaswani. 2018.
\newblock \href {https://arxiv.org/abs/1803.02155} {Self-attention with
  relative position representations}.
\newblock In \emph{Proceedings of the 2018 Conference of the North American
  Chapter of the Association for Computational Linguistics: Human Language
  Technologies, Volume 2 (Short Papers)}, pages 464--468.

\bibitem[{Su et~al.(2019{\natexlab{a}})Su, Shen, Zhang, Sun, Hu, Niu, and
  Zhou}]{su2019improving}
Hui Su, Xiaoyu Shen, Rongzhi Zhang, Fei Sun, Pengwei Hu, Cheng Niu, and Jie
  Zhou. 2019{\natexlab{a}}.
\newblock \href {https://arxiv.org/abs/1906.07004} {Improving multi-turn
  dialogue modelling with utterance rewriter}.
\newblock In \emph{Proceedings of the 57th Annual Meeting of the Association
  for Computational Linguistics}, pages 22--31.

\bibitem[{Su et~al.(2019{\natexlab{b}})Su, Shen, Zhang, Sun, Hu, Niu, and
  Zhou}]{DBLP:rewrite_transformer}
Hui Su, Xiaoyu Shen, Rongzhi Zhang, Fei Sun, Pengwei Hu, Cheng Niu, and Jie
  Zhou. 2019{\natexlab{b}}.
\newblock \href {https://doi.org/10.18653/v1/P19-1003} {Improving multi-turn
  dialogue modelling with utterance {R}e{W}riter}.
\newblock In \emph{Proceedings of the 57th Annual Meeting of the Association
  for Computational Linguistics}, pages 22--31, Florence, Italy. Association
  for Computational Linguistics.

\bibitem[{Vakulenko et~al.(2021)Vakulenko, Longpre, Tu, and
  Anantha}]{DBLP:pipeline_cqa}
Svitlana Vakulenko, Shayne Longpre, Zhucheng Tu, and Raviteja Anantha. 2021.
\newblock \href {https://arxiv.org/abs/2004.14652} {Question rewriting for
  conversational question answering}.
\newblock In \emph{WSDM}.

\bibitem[{Vaswani et~al.(2017)Vaswani, Shazeer, Parmar, Uszkoreit, Jones,
  Gomez, Kaiser, and Polosukhin}]{transformer}
Ashish Vaswani, Noam Shazeer, Niki Parmar, Jakob Uszkoreit, Llion Jones,
  Aidan~N Gomez, \L~ukasz Kaiser, and Illia Polosukhin. 2017.
\newblock \href
  {https://proceedings.neurips.cc/paper/2017/file/3f5ee243547dee91fbd053c1c4a845aa-Paper.pdf}
  {Attention is all you need}.
\newblock In \emph{Advances in Neural Information Processing Systems},
  volume~30. Curran Associates, Inc.

\bibitem[{Wang et~al.(2020)Wang, Shin, Liu, Polozov, and
  Richardson}]{wang2020rat}
Bailin Wang, Richard Shin, Xiaodong Liu, Oleksandr Polozov, and Matthew
  Richardson. 2020.
\newblock \href {https://doi.org/10.18653/v1/2020.acl-main.677} {{RAT-SQL}:
  Relation-aware schema encoding and linking for text-to-{SQL} parsers}.
\newblock In \emph{Proceedings of the 58th Annual Meeting of the Association
  for Computational Linguistics}, pages 7567--7578, Online. Association for
  Computational Linguistics.

\bibitem[{Wang et~al.(2021)Wang, Ling, Zhou, and Hu}]{istsql}
Run-Ze Wang, Zhen-Hua Ling, Jingbo Zhou, and Yu~Hu. 2021.
\newblock \href {https://arxiv.org/abs/2012.04995} {Tracking interaction states
  for multi-turn text-to-sql semantic parsing}.
\newblock In \emph{Proceedings of the AAAI Conference on Artificial
  Intelligence}.

\bibitem[{Yang et~al.(2022)Yang, Ma, Dong, Huang, Huang, Yin, Zhang, Yang, Li,
  and Wei}]{ganlm}
Jian Yang, Shuming Ma, Li~Dong, Shaohan Huang, Haoyang Huang, Yuwei Yin,
  Dongdong Zhang, Liqun Yang, Zhoujun Li, and Furu Wei. 2022.
\newblock \href {https://doi.org/10.48550/arXiv.2212.10218} {Ganlm:
  Encoder-decoder pre-training with an auxiliary discriminator}.
\newblock \emph{CoRR}, abs/2212.10218.

\bibitem[{Yang et~al.(2020)Yang, Ma, Zhang, Wu, Li, and Zhou}]{alm}
Jian Yang, Shuming Ma, Dongdong Zhang, Shuangzhi Wu, Zhoujun Li, and Ming Zhou.
  2020.
\newblock \href {https://ojs.aaai.org/index.php/AAAI/article/view/6480}
  {Alternating language modeling for cross-lingual pre-training}.
\newblock In \emph{The Thirty-Fourth {AAAI} Conference on Artificial
  Intelligence, {AAAI} 2020, The Thirty-Second Innovative Applications of
  Artificial Intelligence Conference, {IAAI} 2020, The Tenth {AAAI} Symposium
  on Educational Advances in Artificial Intelligence, {EAAI} 2020, New York,
  NY, USA, February 7-12, 2020}, pages 9386--9393. {AAAI} Press.

\bibitem[{Yang et~al.(2019)Yang, Wu, Yang, Xu, and Li}]{low_resource_template}
Ze~Yang, Wei Wu, Jian Yang, Can Xu, and Zhoujun Li. 2019.
\newblock \href {https://doi.org/10.18653/v1/D19-1197} {Low-resource response
  generation with template prior}.
\newblock In \emph{Proceedings of the 2019 Conference on Empirical Methods in
  Natural Language Processing and the 9th International Joint Conference on
  Natural Language Processing, {EMNLP-IJCNLP} 2019, Hong Kong, China, November
  3-7, 2019}, pages 1886--1897. Association for Computational Linguistics.

\bibitem[{Yin and Neubig(2018)}]{yin2018tranx}
Pengcheng Yin and Graham Neubig. 2018.
\newblock \href {https://doi.org/10.18653/v1/D18-2002} {{TRANX}: A
  transition-based neural abstract syntax parser for semantic parsing and code
  generation}.
\newblock In \emph{Proceedings of the 2018 Conference on Empirical Methods in
  Natural Language Processing: System Demonstrations}, pages 7--12, Brussels,
  Belgium. Association for Computational Linguistics.

\bibitem[{Yu et~al.(2020)Yu, Liu, Yang, Xiong, Bennett, Gao, and
  Liu}]{yu_Few-Shot_rewrite}
Shi Yu, Jiahua Liu, Jingqin Yang, Chenyan Xiong, Paul~N. Bennett, Jianfeng Gao,
  and Zhiyuan Liu. 2020.
\newblock Few-shot generative conversational query rewriting.
\newblock In \emph{SIGIR}.

\bibitem[{Yu et~al.(2021{\natexlab{a}})Yu, Wu, Lin, Wang, Tan, Yang, Radev,
  Socher, and Xiong}]{yu2020grappa}
Tao Yu, Chien-Sheng Wu, Xi~Victoria Lin, Bailin Wang, Yi~Chern Tan, Xinyi Yang,
  Dragomir Radev, Richard Socher, and Caiming Xiong. 2021{\natexlab{a}}.
\newblock \href {https://openreview.net/forum?id=kyaIeYj4zZ} {{G}ra{PP}a:
  grammar-augmented pre-training for table semantic parsing}.
\newblock In \emph{International Conference on Learning Representations}.

\bibitem[{Yu et~al.(2019{\natexlab{a}})Yu, Zhang, Er, Li, Xue, Pang, Lin, Tan,
  Shi, Li, Jiang, Yasunaga, Shim, Chen, Fabbri, Li, Chen, Zhang, Dixit, Zhang,
  Xiong, Socher, Lasecki, and Radev}]{yu2020cosql}
Tao Yu, Rui Zhang, Heyang Er, Suyi Li, Eric Xue, Bo~Pang, Xi~Victoria Lin,
  Yi~Chern Tan, Tianze Shi, Zihan Li, Youxuan Jiang, Michihiro Yasunaga,
  Sungrok Shim, Tao Chen, Alexander Fabbri, Zifan Li, Luyao Chen, Yuwen Zhang,
  Shreya Dixit, Vincent Zhang, Caiming Xiong, Richard Socher, Walter Lasecki,
  and Dragomir Radev. 2019{\natexlab{a}}.
\newblock \href {https://doi.org/10.18653/v1/D19-1204} {{C}o{SQL}: A
  conversational text-to-{SQL} challenge towards cross-domain natural language
  interfaces to databases}.
\newblock In \emph{Proceedings of the 2019 Conference on Empirical Methods in
  Natural Language Processing and the 9th International Joint Conference on
  Natural Language Processing (EMNLP-IJCNLP)}, pages 1962--1979, Hong Kong,
  China. Association for Computational Linguistics.

\bibitem[{Yu et~al.(2021{\natexlab{b}})Yu, Zhang, Polozov, Meek, and
  Awadallah}]{yu2021score}
Tao Yu, Rui Zhang, Alex Polozov, Christopher Meek, and Ahmed~Hassan Awadallah.
  2021{\natexlab{b}}.
\newblock \href {https://openreview.net/forum?id=oyZxhRI2RiE} {Score:
  Pre-training for context representation in conversational semantic parsing}.
\newblock In \emph{International Conference on Learning Representations}.

\bibitem[{Yu et~al.(2018)Yu, Zhang, Yang, Yasunaga, Wang, Li, Ma, Li, Yao,
  Roman, Zhang, and Radev}]{yu2018spider}
Tao Yu, Rui Zhang, Kai Yang, Michihiro Yasunaga, Dongxu Wang, Zifan Li, James
  Ma, Irene Li, Qingning Yao, Shanelle Roman, Zilin Zhang, and Dragomir Radev.
  2018.
\newblock \href {https://doi.org/10.18653/v1/D18-1425} {{S}pider: A large-scale
  human-labeled dataset for complex and cross-domain semantic parsing and
  text-to-{SQL} task}.
\newblock In \emph{Proceedings of the 2018 Conference on Empirical Methods in
  Natural Language Processing}, pages 3911--3921, Brussels, Belgium.
  Association for Computational Linguistics.

\bibitem[{Yu et~al.(2019{\natexlab{b}})Yu, Zhang, Yasunaga, Tan, Lin, Li, Er,
  Li, Pang, Chen, Ji, Dixit, Proctor, Shim, Kraft, Zhang, Xiong, Socher, and
  Radev}]{yu2019sparc}
Tao Yu, Rui Zhang, Michihiro Yasunaga, Yi~Chern Tan, Xi~Victoria Lin, Suyi Li,
  Heyang Er, Irene Li, Bo~Pang, Tao Chen, Emily Ji, Shreya Dixit, David
  Proctor, Sungrok Shim, Jonathan Kraft, Vincent Zhang, Caiming Xiong, Richard
  Socher, and Dragomir Radev. 2019{\natexlab{b}}.
\newblock \href {https://doi.org/10.18653/v1/P19-1443} {{SP}ar{C}: Cross-domain
  semantic parsing in context}.
\newblock In \emph{Proceedings of the 57th Annual Meeting of the Association
  for Computational Linguistics}, pages 4511--4523, Florence, Italy.
  Association for Computational Linguistics.

\bibitem[{Zhang et~al.(2019)Zhang, Yu, Er, Shim, Xue, Lin, Shi, Xiong, Socher,
  and Radev}]{zhang2019editing}
Rui Zhang, Tao Yu, Heyang Er, Sungrok Shim, Eric Xue, Xi~Victoria Lin, Tianze
  Shi, Caiming Xiong, Richard Socher, and Dragomir Radev. 2019.
\newblock \href {https://doi.org/10.18653/v1/D19-1537} {Editing-based {SQL}
  query generation for cross-domain context-dependent questions}.
\newblock In \emph{Proceedings of the 2019 Conference on Empirical Methods in
  Natural Language Processing and the 9th International Joint Conference on
  Natural Language Processing (EMNLP-IJCNLP)}, pages 5338--5349, Hong Kong,
  China. Association for Computational Linguistics.

\bibitem[{Zheng et~al.(2022)Zheng, Wang, Dong, Wang, and Li}]{hie-sql}
Yanzhao Zheng, Haibin Wang, Baohua Dong, Xingjun Wang, and Changshan Li. 2022.
\newblock \href {https://arxiv.org/abs/2203.07376} {{HIE-SQL}: History
  information enhanced network for context-dependent text-to-sql semantic
  parsing}.
\newblock In \emph{Findings of the Association for Computational Linguistics
  2022}, Online. Association for Computational Linguistics.

\bibitem[{Zhong et~al.(2020)Zhong, Lewis, Wang, and
  Zettlemoyer}]{zhong-etal-2020-grounded}
Victor Zhong, Mike Lewis, Sida~I. Wang, and Luke Zettlemoyer. 2020.
\newblock \href {https://doi.org/10.18653/v1/2020.emnlp-main.558} {Grounded
  adaptation for zero-shot executable semantic parsing}.
\newblock In \emph{Proceedings of the 2020 Conference on Empirical Methods in
  Natural Language Processing (EMNLP)}, pages 6869--6882, Online. Association
  for Computational Linguistics.

\bibitem[{Zhong et~al.(2017)Zhong, Xiong, and Socher}]{zhongSeq2SQL2017}
Victor Zhong, Caiming Xiong, and Richard Socher. 2017.
\newblock \href {https://arxiv.org/abs/1709.00103} {Seq2sql: Generating
  structured queries from natural language using reinforcement learning}.
\newblock \emph{CoRR}, abs/1709.00103.

\end{thebibliography}
\bibliographystyle{acl_natbib}

\clearpage
\appendix

\section{Example of Bi-directional Rewriting Matrix}
\label{appendix_b}
\begin{figure}[!h]
	  \centering
	  \setlength{\abovecaptionskip}{8pt}
	  \setlength{\belowcaptionskip}{-2pt}
	   \includegraphics[width=7.7cm]{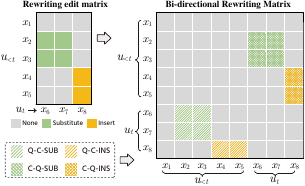}
	   \caption{An example of bi-directional rewriting matrix for two-stream matrix relation encoder.}\label{example2}
\end{figure}

The method to extend the uni-directional rewriting matrix to the bi-directional rewriting matrix is illustrated in Figure~\ref{example2}.
When encoding rewriting matrix in two-stream encoder, question $u_t$ is concatenated with context $u_{<t}$. Given question context $u_{<t}=\{x_1,x_2,...,x_5\}$ and current question $u_t=\{x_6,x_7,x_8\}$, we first generate uni-directional rewriting matrix, which contain three type relations: “\texttt{None}”, “\texttt{Substitute}”, and “\texttt{Insert}”. Then we extend relations to bi-directional: “\texttt{substitute}” is extended to “\textsc{{Q-C-Sub}}” and “\textsc{{C-Q-Sub}}”, “\texttt{insert}” is extended to “\textsc{{Q-C-Ins}}” and “\textsc{{C-Q-Ins}}”.\vspace{-5pt}
\section{Experimental Details of Query Rewriting Model}
\label{appendix_a}
\subsection{Dataset}
We train and evaluate the QR model on the \textsc{Canard} dataset~\cite{canard} which consists of $31$K, $3$K, and $5$K QA pairs for training, development, and test sets, respectively. The semantic-incomplete questions in \textsc{Canard} are generated by rewriting a subset of the original questions in QuAC~\cite{quac}.

\subsection{Implementation Details}
We adopt a pre-trained T5-base model to initialize our QR model, the learning rate is $1e$-$4$, the batch size is $256$, the number of training epochs is $20$ and we use AdamW optimizer with linear warmup scheduler. Model are trained with $8$ NVIDIA V100 GPU cards.

\subsection{Metrics And Results}
We employ automatic metrics $\mathbf{ROUGE}$ to evaluate the QR model, where $\mathbf{ROUGE}_{n} (\mathbf{R}_{m})$ measures the $n$-gram overlapping between the rewritten questions and the
golden ones, $\mathbf{ROUGE}_{L}(\mathbf{R}_{L})$ measures the longest matching sequence between them. As shown in Table~\ref{canard_result}, our QR model gets better results against all baselines on the \textsc{Canard} dataset. \vspace{-5pt}

\begin{table}[!h]
\centering
\addtolength{\tabcolsep}{1.5mm}
\resizebox{0.45\textwidth}{!}{
\begin{tabular}{l|ccc}  
\toprule[1.0pt]
\textbf{Models}&$\mathbf{R_{1}}$&$\mathbf{R_{2}}$&$\mathbf{R_{L}}$ \\
\midrule[0.3pt]
Copy$^{\dag}$ &72.7&54.9&68.5 \\
Pronoun Pub$^{\dag}$ &73.1&63.7&73.9 \\
L-Ptr-GEN$^{\dag}$ & 78.9 &62.9&74.9 \\
RUN$^{\dag}$ & 79.1&61.2&74.7 \\
\textbf{T5-base (Ours)}&\textbf{81.3}&\textbf{70.1}&\textbf{78.4} \\
\bottomrule[1.0pt]
\end{tabular}
}
\caption{Evaluation results of the QR models on \textsc{Canard} dataset.${\dag}$: Results  are from \cite{liu2020incomplete}} 
\label{canard_result}
\end{table}

\subsection{Rewriting Matrix Evaluation}
To verify that rewriting edit matrix holds the key rewritten information from rewritten question,
we compare the restored rewritten question from rewriting matrix (denoted as \textsc{Res-RQ}) with rewritten question generated by QR model (denoted as \textsc{Gen-RQ}) and evaluate it with $\mathbf{ROUGE}$. As shown in Table~\ref{tab_restore_rewriting_matrix}, we take the \textsc{Gen-RQ} as the target and evaluate the difference between the target and \textsc{Res-RQ}. On both datasets, \textsc{Res-RQ} and \textsc{Gen-RQ} maintain a high level of consistency. This indicates that the rewriting matrix effectively stores the key information of question rewriting.
\begin{table}[h]
\centering
\setlength{\abovecaptionskip}{5pt}
\setlength{\belowcaptionskip}{-7pt}
\resizebox{0.485\textwidth}{!}{
\begin{tabular}{l|cccccc} 
\toprule[1.0pt]
\multirow{2}{*}{\!\!\textbf{Datasets}}&\multicolumn{3}{c}{\textbf{Train}}&\multicolumn{3}{c}{\textbf{Dev}}\\
&$\mathbf{R_{1}}$&$\mathbf{R_{2}}$&$\mathbf{R_{L}}$&$\mathbf{R_{1}}$&$\mathbf{R_{2}}$&$\mathbf{R_{L}}$\\
\midrule[0.3pt]
\!\!\textsc{Sparc}&91.7&81.6&90.0&91.8&82.0&90.3\\
\!\!\textsc{CoSQL}&89.0&81.5&87.9&88.4&80.6&87.2\\
\bottomrule[1.0pt]
\end{tabular}
}
\caption{Evaluation results of restoring rewritten question from rewriting matrix.}
\label{tab_restore_rewriting_matrix}
\end{table}

\end{document}